# An Interesting Uncertainty-based Combinatoric Problem in Spare Parts Forecasting: The FRED System

John B. Bacon
Senior Systems Analyst
Xerox Corporation 0207-01D
Webster, NY 14580

**Abstract:**
   The domain of spare parts forecasting is examined, and is found to present unique uncertainty-based problems in the architectural design of a knowledge-based system. A mixture of different uncertainty paradigms is required for the solution, with an intriguing combinatoric problem arising from an uncertain choice of inference engines. Thus, uncertainty in the system is manifested in two different meta-levels. The different uncertainty paradigms and meta-levels must be integrated into a functioning whole. FRED is an example of a difficult real-world domain to which no existing uncertainty approach is completely appropriate. This paper discusses the architecture of FRED, highlighting: the points of uncertainty and other interesting features of the domain, the specific implications of those features on the system design (including the combinatoric explosions), their current implementation & future plans, and other problems and issues with the architecture.

## Introduction:

   The Failure Rate Estimating Device (FRED) is being developed as a robust design tool for use by engineers and designers within Xerox Corporation. It is an intelligent decision support system which: identifies potential spare parts in a new electromechanical assembly, isolates likely failure modes and rates, suggests corrective redesigns, and finally, calculates economic levels of assembly and stock quantities for all spared parts and higher assemblies. Such an endeavor, in which the system is asked to predict the future failures of new designs, is necessarily a highly uncertain domain. To Xerox (& to many other mechanical equipment manufacturers), it is a critical problem centered in human expertise, which is worth tens of **millions** of dollars annually in potential cost avoidance.

   A copier may have in excess of 5000 unique piece-parts, many of which will have to be replaced during the life of the machine. The goal of the spares forecasting effort is to predict and to optimize copier reliability, service cost, repair time, and inventory levels for all necessary part replacements. The parts are arranged (in a pure tree structure) into **many** higher levels of assembly, any of which may be spared as a whole, as part of an even higher assembly, or BOTH. For instance, in a spare tire, the valve cap is usually available to the consumer as a component, as part of a spare valve assembly, and as part of a complete tire change.

   There are **different** methods of estimating the combined failure rate of a high-level assembly from the individual failure rates of its parts. All later economic analyses derive from this estimate. The choice of method is dependent upon each failure mode's classification as either a wearout or a random mode of failure. When this choice is uncertain (as it is for many of these unique parts), a combinatoric explosion of failure rate possibilities can be generated, even if the exact individual rates of failure are known.(They are not known exactly: a **range** of possibilities is considered, and treated by the expert as a three-valued set. ( high , low, and best estimates.)

## The Problem Space & typical expert solution:
### 1)Spare Candidacy:

   The experts in this domain pursue the following strategy. First, they work very shallowly, breadth first, doing an opportunistic search over the part and assembly tree for familiar classes of typical spares and non-spares. Along with this initial classification, they do an approximate total of weights and costs, and then mark the places in the assembly where it is either too heavy, costly, or complex to spare. Any known class of spare (eg, "BELT" or "BEARING", etc) and **all** of its higher assemblies up to the marked limits are marked as potential spare candidates. Note that there is no ONE sparing level choice at this point: it is common to have multiple sparing levels within a single assembly (as in the valve-cap example, above).

   The expert will refine his decisions first by seeking the inseparable points in the assembly. Like the weight and cost totals, this too limits the search space. This eliminates all potential spare candidates that are lower in the tree than the inseparable points. The expert may then refine his earlier estimates of weight and cost by a more precise addition of all subcomponent weights & costs, and thus have to change some earlier decisions about spares candidacy. Through this point in the process, the expert is simply trying to identify the major points of failure in the assembly, and a complete range of parts replacement possibilities. Whenever the expert identifies a potential reliability problem area, he explores possible corrective design changes with the subsystem engineer. This may happen at any point in the process.

**Failure Modes & Rates:**
The expert then begins to identify the specific failure modes and the associated rates for each piece-part in the assembly. (These are the end or leaf nodes in the part and assembly tree. Individual parts can have many modes of failure.) Each failure mode is classified as either RANDOM or WEAROUT. For the known classes of spare candidates, the expert has pre-compiled statistical data on earlier designs, from which he can extract a tight **range** of estimates for the individual and combined failure rates of each part. He speaks of this range as a **set** of three values: a high estimate, a low estimate, and a best estimate. The expert also makes a note of the impact of each failure mode on the overall performance of the copier. (eg: until it can be fixed, will this failure mode shut down the process completely, or just slightly degrade the copy quality?). On the average, over two thirds of all sparing choices have been firmly decided at this point. When he has been *unable* to classify the part, the expert then resorts to much deeper reasoning.

**Deeper reasoning:**
   The expert looks for key indicators of failure, checking for new manufacturing processes, any of several materials, geometric features such as sharp corners, and most importantly, the usage of the part, both in terms of actual load types & levels, and on functional use in the machine.§ The expert will also carefully examine laboratory test data on wear and failure of new designs. From all of these pieces of partial evidence, he then constructs an (uncertain) scenario of the set of failure modes for each part, and assigns an even more uncertain failure rate to each mode. He will also increase or decrease the estimate of failure rate gleaned earlier from the known class data, based upon additional knowledge about usage or any other distinguishing feature. We intend to use Bayesian inferencing (Schmitt,1969) to model this behavior in FRED.
   Note that even the random failure rate is a **measure** of uncertainty: it is a representation of the probability that any given part will fail within a certain number of copy cycles. It is normalized into a guess as to how many part replacements will be necessary per million copies made by the **entire fleet** of copiers using this particular design. At this point, all failure modes and rates have been identified for each piece part. Problems begin to compound as one tries to find the **total** failure rate of each part, and of all higher assemblies.

**Combining rates:**
This next step is the root of the major uncertainty difficulties in FRED. RANDOM failure rates are **added** for each part, since any random failure may occur anywhere within the life of the copier. However, WEAROUT rates for any part or assembly are grouped, and the **maximum** rate of all members of this set is taken as the sole rate.† A total fleet failure rate is maintained by **adding** the maximum wearout rate to the combined random rates at each point in the part & assembly tree. ‡
   Random rates represent a statistical probability of failure of any given part, whereas wearout rates represent an estimated life. If a particular failure mode cannot be assigned with certainty to be definitely random or wearout, the combinatorics become difficult. Our experts then apply a higher level of reasoning to the problem, in a effort to avoid costly extended life-cycle testing of the new design. (In an expert system, this implies a jump to a higher meta-level!)

---

§ e.g.: Parts associated with the main drive train are historically more susceptible to failure than are other gears, pulleys, etc., due to the high historical likelihood that tolerance stack-ups & minor design errors will throw the actual loads well outside of the design envelope.

† The point is that *failure will occur at a predictable time*. Thus, if there are several wearout modes in a part or assembly, the actual rate is the MAXIMUM wearout rate: the other wearout rates are ignored, since the part will have failed before any of these modes can occur.

‡ Over a population of fifty thousand copiers, Xerox will have to spare for **both** types of failure of each part.

Faced with an uncertainty about the actual failure rate, the experts resort to goal-driven reasoning. They work backwards from the possible outcomes (for level of repair) of the pending economic analysis of the assembly. There are a variety of real-world pressures which also influence the decision, and these are consulted to see if any of them would dominate the level of repair, stocking echelon, and stocking rate decisions.

If all else fails, the expert resorts to an elementary decision-theoretic approach. He makes his decision to pursue further experimentation by looking at whether the cost of isolating the exact failure mode will outweigh the cost to the company of buying and stocking spares under the wrong random/wearout assumption. This of course requires economic calculation of all possibilities for spare candidacy. The problem is compounded by the fact that the experts consider a **distribution** of failure rate possibilities for each mode of every part. The experts exhibit fuzzy reasoning in maintaining a **three-element range set** as a representation of the possibility distribution.

FIGURE 1: **Part and assembly spares tree.** *This is a FRED representation of a typical part and assembly tree. Spare parts candidates are flagged as the black nodes, known or presumed non-spares are flagged with gray borders, and currently unknown/undecided parts would have no borders. FRED has noticed that* **Bkt-Assy-vcf** *contains a* **Rivet**, *which makes it inseparable, so it* **must** *be a spare to allow replacement of the* **Spring-clip**, *which is a known type of failure-prone part. There are several spare candidates in* **Shaft-Assy-VCF**, *made inseparable by the* **Spring-Pin:** *if we don't know whether they are random or wearout, how can we calculate the total failure rate of* **Shaft-Assy-VCF**?

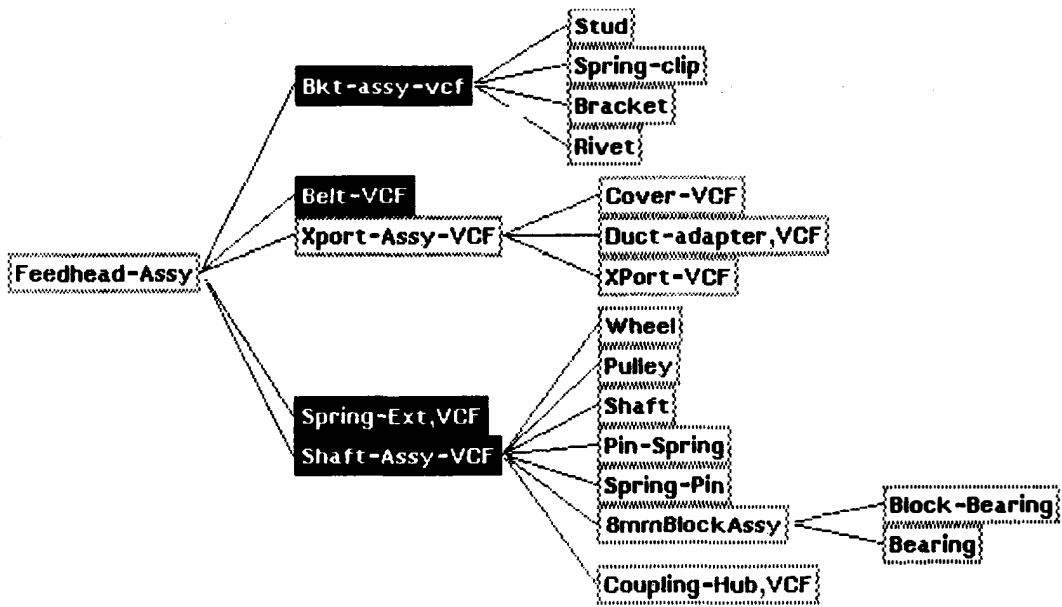

**Entry of additional modes and rates:**
At the assembly level, <u>additional failure modes</u> are usually introduced relating to the assembly itself, independent of any particular part within it. Misalignment is the common failure mode of the assembly, and it can have **both** random (abuse) **and** wearout (normal use) modes. Whether or not misalignment implies a spare candidacy is dependent upon relationships between the cost or weight of the assembly, the time to repair it in the field, the required precision of the alignment, and the required skills & tools of the service representative to effect the repair. The expert makes judgment calls on each "assembly mode of failure", and if required, adds the failure rate into the rates for the assembly. Thus, failure rate at each branch point of the tree cannot be modelled as just the combination of rates of all of its subassemblies

**Level of Repair:**
The expert seeks to <u>group</u> individual piece parts into whole <u>modules or kits</u>, based upon the wearout rates & on ease of field replacement. The purpose here is to trigger preventative maintenance of all parts with similar wearout rates whenever the first of this group wears out, for the others' failure will obviously be imminent. (Note that this procedure applies to wearout failure modes **only**, and is thus sensitive to the uncertainty in mode type classification!). Next, he performs an economic comparison of the relative costs associated with copier disassembly & individual part replacement versus simple replacement of the entire failed module. The trade-off here is in material cost versus labor cost and lost revenue from the copier "down time". In a reasonably simple economic algorithm, he finally assigns the final spare decisions on each spare candidate node. It is possible that a fallible part may not be spared individually, or that it may be multiply spared, as in the tire valve-cap example. Each sparing decision reduces the predicted residual failure rate of higher assemblies, and preempts any wearout modes of lower assemblies. During this and the subsequent phase of the solution, the expert highlights those sensitive areas where his decision could be easily swayed either by improved data or improved part performance. Based upon this sensitivity analysis, he makes recommendations to the subsystem engineer for further experimentation or redesign.

### Stocking Echelon & Stocking Level:
The final decision is that of picking a **place** to locate the spare parts. (Stocking Echelon) There are four possibilities: ranging from the service representative's car supplies to the factory floor (ie, make on demand). Here too, the expert may choose to locate the same spare in several locations, to have an average and a readily available back-up supply. The expert will also pick the quantity to be stored (Stocking Level) at each location, based upon estimated failure rate, on replacement part cost. (e.g.: It is preferable to put even high-failure rate, expensive items (such as integrated circuit boards) in centralized locations, and to ship them via courier, rather than to have to "fill the pipeline" with them.) This decision is also dependent upon the expected monthly copy volume, (this scales how many actual failures will be observed) the **severity of problem caused by the failure,** and on the age of the fleet. (e.g.: when will the fleet start approaching the wearout life of each part?).

# Domain Analysis:
From the preceding discussion, the following can be deduced about uncertainty in the spares forecasting process:
- Failure Rate is in itself a measure of uncertainty.
  - Random Failure rate is a **direct** representation of the certainty of failure.
  - Complexity of the decision process sometimes forces the expert to roll his uncertainty about **mode** into an increased failure **rate**.
- The process of *estimating* of the rate is itself uncertain, and is typically a Bayesian process.
  - Failure modes and rates are often estimated from partial and conflicting evidence.
  - Upshifting and downshifting of rate is affected by the addition of more data.
- The experts propagate uncertain estimates as fuzzy sets: not as continuous distributions.
- The type of failure **mode** is sometimes uncertain, especially when working with observed laboratory failures.
  - *Propagation of rates through the part & assembly tree differs, depending on random vs. wearout mode.*
- Belief Revision & Truth Maintenance are absolutely necessary to the solution:
  - Many spare candidacy decisions are non-monotonic.
  - The data will change in response to expert suggestions, "un-doing" trouble-spots.
  - The principle of least commitment is evident in the solution.
- Three distinct optimization searches (Level of assembly, stocking echelon, stocking rates) on multiple alternatives are performed, based upon uncertain inputs (failure rates, modes, repair time). A sensitivity analysis on the impact of uncertain decisions is the basis for the expert's advice on the design changes. Thus:
  - There is a need to maintain multiple, alternate *hypotheses* in the Level of Repair economic analysis of the assembly.
  - There can be multiple, simultaneous *solutions* to the sparing of any one part.
  - There can be multiple, simultaneous solutions to the *stocking location and level* of any one spare.
- Several design alternatives may be maintained concurrently during the design process.

**FIGURE 2:** *These issues are the driving forces which will determine the architecture of the inference engine for FRED. The diagram pictorially represents the entry and flow of uncertainty in the spares decision-making process.*

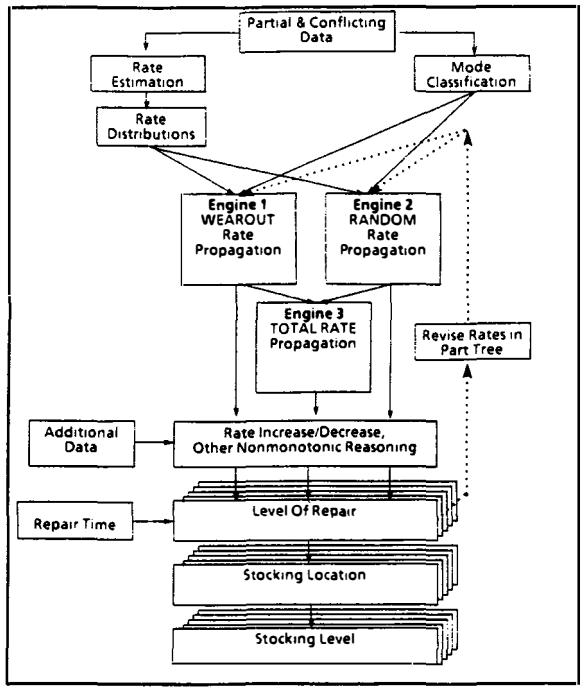

**FIGURE 3:** *Effects upon the data of the uncertainty in the choice of inference engine.*

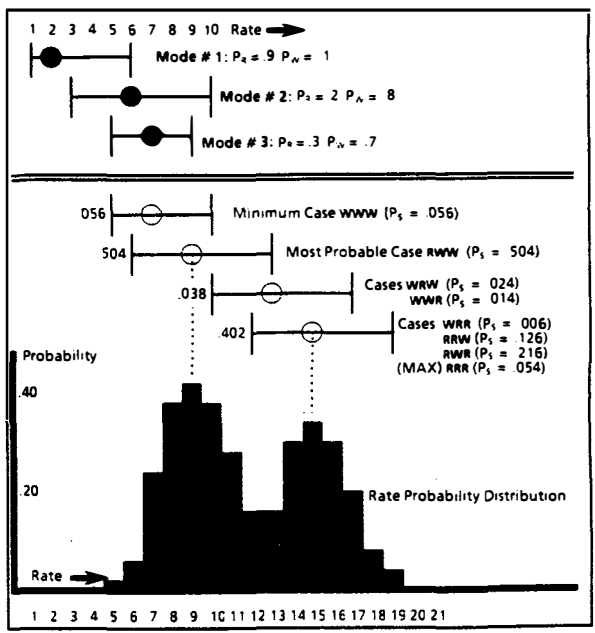

## Base Architecture:

FRED is built in the LOOPS environment, with each new part or assembly being represented by a single ("prototype") instance of a unique object class. This class is defined by attaching to it multiple selected parent classes, "Mixins", and (eventually) Rule Objects, to provide general methods and constraints applicable to the new part. Key data is contained in instance variables, some of which will be maintained as "Active Values", capable of triggering rule sets whenever they are updated. Most of the default reasoning is already fired immediately through use of the methods embedded within the main program control structure. The rule sets will handle the more complicated deep reasoning: there will be both forward chaining and backward chaining rule sets. Reasoning hitory lists will be maintained for each part and assembly, and will be used for the explanation utility.

## Uncertainty Problems in the architecture:

Consider the problem of determining the combined failure rate of 3 distinct failures. (Either 3 types of failure of a single part, or else 3 potentially failing parts in one assembly.) Each failure rate carries a best estimate (80% weight), an upper and lower extreme on the range of possibilities (10% weight each), and a probability Pr and Pw that the random or wearout failure mode is correct. (Pw + Pr = 1).

Each of the following scenarios can be immediately constructed,(for any number of combined rates) and the probabilities of each assigned by multiplying the probabilities of the mode assignments (random vs. wearout).

1) A **maximum** possible failure scenario can be found by assuming that all modes are **Random**. Then, all of the rates must be summed. The equivalent 10% upper limit of the estimation range is the maximum difference between the upper limit and best estimate of the set of all parts, plus the sum of the other four <u>best</u> estimates. (Similarly for the lower limit of the combination estimate)

2) A **most probable** rate scenario can be computed by picking the most probable mode of each of the contained parts, and then combining accordingly.

3) A **minimum** rate scenario can be found by assuming that all of the rates are indeed **Wearout**. The best estimate rate for only one part will apply, and all other rates discarded. (Note that the upper limit will be the **maximum** of the set of all upper limits, and the lower limit will be the **maximum** of all lower limits.)

Note that each scenario yields a three-element fuzzy set (range). The weightings of each element actually represent <u>the likelihood of each rate being accurate in a Bayesian combination with the probability that the entire scenario is correct at all</u>. Probabilistically, there are $2^n$ possible scenarios for each combination of n failure modes, although $n+1$ will always yield the same rate and range estimates (The all random, or maximum rate case, plus each of the cases where exactly one mode is a wearout.) Our experts persist with their 80%, 10%, and 10% weightings of certainty for best, high, and low estimates. Unfortunately, the combinations occasionally give a curve for which such a combined weighting scheme is ambiguous, as can be seen in the histogram of the 3 mode example above. Note that in the example, the most probable scenario **RWW** is only marginally more likely than (and might have been, under slightly different conditions, **less** than) the next most likely case, which represents the $n+1$ cases equal to the maximum **RRR** case. (The scenario labelled **RWW** stands for **R**andom, **W**earout, **W**earout. Similar abbreviations apply for all scenarios. The rates to be combined are at the top of the diagram, with black centers. The rate combination possibilities are below with white centers. **Ps** is the total probability that each scenario is true.)

Note also that the combinatoric explosion arises over the calculation of **rates** and **modes**: each rate/mode scenario must **then** drive the final optimization routines which pick the preferred sparing points and stocking echelons for each identified spare, before the **sensitivity** of the mode-classification decision can be calculated. The final economics are **not** always quickly deducible from the raw rate data. They are calculated in an existing, algorithmic mainframe system which is strongly founded in decision theory. FRED's job is to provide the inputs to this system, yet FRED must have a heuristic understanding of the possible outcomes if it is to efficiently prune the search space. The explosion of the number of mutually exclusive potential failure scenarios makes it impossible to apply the hierarchical reasoning methods espoused by Levitt (Levitt 1986). It is, in fact, the ability to **heuristically** understand the ultimate economic & service quality sensitivities to the raw failure data that allows our **human** experts to work efficiently in this highly complex combinatoric problem area. Their heuristics, however, cover a wealth of worldly experience, and even extend to common sense/common knowledge reasoning, which is beyond the scope of present technology.

Through its heuristic understanding of the optimization principles, FRED must present the user with the **sensitivity** of the final copier reliability (and of the <u>economic</u> outcome), to the assumptions used in the decision.

The following are examples of factors which can preclude the need for a detailed analysis:

**1)** If the rate is small enough (eg: the individual average machine life is less than the estimated rate of failure), then don't stock spares at all, and either cannibalize an older machine, or special-order the part as required if and when it does fail.

**2)** If the part is cheap & small enough, then order the entire quantity needed for the life of the fleet in a bulk order, (to save on manufacturing tooling set-up & shipping/distribution costs) and then store it at the service branches.

**3)** If the assembly is expensive and/or large enough, then the assembly will only be stored in small numbers in one of the two most centralized locations, and shipped as needed. (This is because design changes can cause obsolescence of expensive inventories of parts!).

These, and several other criteria, can be used to eliminate most combinatoric possibilities when the actual mode and rate data are uncertain. The trick in FRED is to develop <u>heuristic weightings</u> to capture the idea of "expensive <u>enough</u>, heavy <u>enough</u>, large <u>enough</u>, severe <u>enough</u>", etc. to interact with and prune a combinatorically explosive set of equations. These are combined currently in EMYCIN-like certainty factors (Shortliffe & Buchanan, 1975) in the prototype backward chaining rule engine. (We are using this paradigm mainly for expedience to capture the knowledge, until we can develop something better.)

## Temporary "Solution":

It is easy to put some global constraints on costs, weights, sizes, etc., and indeed we do so within FRED. It is even possible to do this with the concept of "failure impact severity". However, it is that large gray area where an assembly may be "mildly expensive", with a "moderately high rate of failure", and have a "pretty severe impact" on the copier performance, and yet not exceed any one criterion to trigger a rule that might cut off a combinatoric explosion. While we seek heuristic weightings, we would like **not** to implement the final solution with numeric scoring functions, such as our current EMYCIN certainty factors. This is because our experts do **not** appear to make their decisions in any quantitative way. We are aware that this point in the decision process is subjective: we do not intend to reduce it to mathematical manipulations, if we can possibly avoid it. Further, the work of Hekerman & Horvitz (Hekerman & Horvitz 1986) would indicate that Certainty Factors are not an appropriate medium for nonmonotonic reasoning.

Though we are constructing the prototype rule engine with heuristic weightings, it is often best to resort to the optimum heuristic machine: the user. FRED currently flags the user with any unresolved combinatoric problems, showing the source(s) of the uncertainty driving the potential problem, and the limits to areas of the part & assembly tree which might be impacted by a wrong assumption.(based upon the point at which a global constraint will be reached under **either** the maximum or minimum failure rate case.) The user would then have three choices: either to declare himself to be certain about his input, to pick the spare level and stocking echelon himself, or to run all possible scenarios. (the situation we sought to avoid!)

Our belief is that there will be so few cases which require resolution of this type that the system will appear to reach "good" decisions on-line and in real time for most users. However, the currently unresolved problem: of **heuristic sensitivity analysis on the inference engine itself to avoid a combinatoric possibility explosion,** remains as an interesting problem, and is the barrier which keeps FRED from achieving true expert status.

## Solution Strategy

Our strategy to ultimately resolve the problem is to **continue** pruning the tree, using deeper knowledge, and in general to make optimum use of **all** other knowledge before considering the actual rates of failure. (An example of the *Principle of Least Commitment* (Stefik, 1981)) The control architecture works breadth-first over the part and assembly tree, delaying the detailed node-by-node numeric propagation of rates until much shallower reasoning has pruned the options because of inseparability, cost, and weight, etc. We will rely heavily upon a **forward chaining** rule engine in the breadth-first architecture. We believe that the mentioned deeper knowledge will be coupled to the rate data, and be centered mostly in the final economic and reliability optimization heuristics. Thus, any unresolved issues left over from the forward-chaining (=Data-Driven) reasoning from the spares identification stage will require Goal-Driven (backward chaining) reasoning to the rate data from the optimization heuristics. That is, we will seek to inquire what rate scenarios would make a difference, using a simple risk analysis as a discriminator:

---

**EXAMPLE: a backward-chaining decision-theoretic heuristic:**

For an assembly whose cost has been determined to be in a certain range, and which cannot be otherwise assigned a stocking echelon by the earlier heuristics, we might know from two optimization heuristics that:

1)If the total failure rate is greater than 1 per million we will stock it at the branch and require a total order of 5000 spares.

2)If less than 1 per million, we will stock it at the regional distribution centers, and keep a total of 1000 spares on hand, ready to ship immediately. The cost differential is approximately 5:1.

The minimum and maximum of range of failure possibilities of the assembly straddle the 1 per million rate, so the decision is not obvious. Thus, the rate data and the economic knowledge are coupled, and we wish to implement backward-chaining heuristics in the controlling meta-knowledge that would examine the rate data, to see if the chance was **substantially** less or more than 20% that the rate exceeded 1 per million. If so, the system could decide one way or the other which way to spare the part, without propagating all rate scenarios through the economic optimization engine.

---

We will continue to build upon the idea of fuzzy set representation of the possibility distribution (the {**max** (0.1), **min**(0.1), **best estimate**(0.8)} weightings of rate), as a mechanism for propagating any uncertain data (rate, repair time, etc.). However, the combinatorics of these are simple to follow only if the failure mode is known to be wearout. Additionally, in the sensitivity analysis, it is far simpler to use sets rather than distributions to track backwards to the main drivers of the sparing decisions. The combination of random failures does not lend itself to the traditional Min & Max operators of classical Fuzzy Set Theory (Negoita, 1985), and we have yet to uncover heuristic algorithms to emulate the expert's insistence upon simple three-element set representations of the **combined** failure rates. The set notation is particularly hard to correlate with probability distributions as typified in Figure 3. We will delay **any** propagation of sets until the forward chaining rules are exhausted, and then chain backwards whenever uncertainty in mode assignment exists.

In short, then, we will be using forward chaining rules, default data inheritance, and breadth-first control to limit the scope of the potential combinatoric explosions, while employing deep reasoning with Bayesian evidence combination on a few otherwise unclassifiable parts. We will then use backward-chaining and simple risk analysis to construct tests to eliminate the need for full-blown economic analyses of many of the combinatoric possibilities. The final combinations and sensitivity analyses will be based in fuzzy set manipulations of three-element sets for every node in the part and assembly tree. These sets will require intermediate possibility distribution manipulations. ( Prade, 1985, & Yager, 1982)

## Conclusion:

The spare parts forecasting domain requires a highly non-monotonic process acting with many types and levels of uncertainty. Effective solution of the problem within FRED, a knowledge-based computer system, requires use of several major types of knowledge & uncertainty representation, combination, and propagation. FRED will seek to <u>heuristically</u> control the combinatoric explosion of possibilities at the **data** level, caused by uncertainty at a much deeper **meta** level.

## Acknowledgements

The author wishes to thank Dr. Tod S. Levitt for his encouragement and many constructive suggestions. This work has been sponsored by Xerox Corporation. The opinions expressed herein are those of the author, and are not necessarily those of his employer.

## References:

Heckerman, D.E, & Horvitz, E.J: *The Myth of Modularity in Rule-based Systems* Proccedings of the 2nd Annual AAAI Workshop on Uncertainty In Artificial Intelligence. (University of Pensylvania Aug 8-10, 1986.)

Levitt, T.S: *Model-Based Probabilistic Situation Inference in Hierarchical Hypothesis Spaces* (in: Kanal L.: **Uncertainty in Artificial Intelligence** (Amsterdam: North-Holland 1986))

Negoita, V.N: **Expert Systems and Fuzzy Systems** (Menlo Park, Ca: Benjamin/Cummings 1985)

Prade, H: *Reasoning with Fuzzy Values* The 15th Int. Symposium on Multiple-Valued Logic (1985: IEEE Computer Society Press) pp 191-197

Schmitt, S.A: **Measuring Uncertainty** An Elementary Introduction to Bayesian Statistics (Reading, Ma: Addison Wesley 1969)

Shortliffe, E.H, &Buchanan, B.G: *A Model of Inexact Reasoning in Medicine* Mathematical Biosciences **23**, pp 351-379  1975

Stefik, M: *Planning with Constraints* Artificial Intelligence **16**, 2 pp 111-140  1981

Yager, R.R: *Generalized Probabilities of Fuzzy Events from Fuzzy Belief Structures* Information Sciences **28**, pp 45-62  1982